\documentclass[10pt,journal,compsoc]{IEEEtran}
\ifCLASSOPTIONcompsoc
  \usepackage[nocompress]{cite}
\else
  \usepackage{cite}
\fi

\ifCLASSINFOpdf

\else

\fi

\usepackage{graphicx}  %insert graph
\usepackage{epstopdf}  %insert eps file
\usepackage{multirow}
\usepackage{array}  %assign the width
\usepackage{color}
\usepackage{colortbl}
\usepackage{booktabs}
\usepackage{amsmath}
\usepackage{amssymb}
\usepackage{amsfonts}  %use hollow body letters
\usepackage{amssymb}
\usepackage{wasysym}
\usepackage{bm}
\usepackage{mathrsfs}  %use curlicue letters
\usepackage[switch]{lineno}  %\linenumbers
\usepackage{algorithm}
\usepackage{algpseudocode}
\usepackage{booktabs}
\usepackage{threeparttable}
\usepackage{url}
\usepackage{xcolor}
\newcommand{\textred}[1]{\textcolor[rgb]{0.84,0.1,0.1}{#1}}
\newcommand{\textgreen}[1]{\textcolor[rgb]{0.314,0.686,0.314}{#1}}
\newcommand{\textgray}[1]{\textcolor[rgb]{0.608,0.608,0.608}{#1}}

\begin{document}
%
% paper title
\title{Equivalent Classification Mapping for Weakly Supervised Temporal Action Localization}

\author{Tao~Zhao, Junwei~Han,~\IEEEmembership{Senior~Member,~IEEE,} Le~Yang, and Dingwen~Zhang~\IEEEmembership{Member,~IEEE}% <-this % stops a space
\IEEEcompsocitemizethanks{\IEEEcompsocthanksitem This work was supported by the Key-Area Research and Development Program of Guangdong Province(2019B010110001), the Research Funds for Interdisciplinary subject NWPU, the China Postdoctoral Support Scheme for Innovative Talents under Grant BX20180236, the National Natural Science Foundation of China under Grants 61876140 and U1801265. (\emph{Corresponding authors: Junwei Han and Dingwen Zhang.})
\IEEEcompsocthanksitem T. Zhao, J. Han, L. Yang, and D. Zhang are with the School of Automation, Northwestern Polytechnical University. D. Zhang is also with the School of Mechanoelectronic Engineering, Xidian University. (e-mails: junweihan2010@gmail.com and zhangdingwen2006yyy@gmail.com).}% <-this % stops an unwanted space
\thanks{}}

% The paper headers
\markboth{IEEE TRANSACTIONS ON PATTERN ANALYSIS AND MACHINE INTELLIGENCE, VOL. -, NO. -, - 2020}%
{Shell \MakeLowercase{\textit{et al.}}: Bare Demo of IEEEtran.cls for Computer Society Journals}

\IEEEtitleabstractindextext{%
\begin{abstract}
	Weakly supervised temporal action localization is a newly emerging yet widely studied topic in recent years. The existing methods can be categorized into two localization-by-classification pipelines, i.e., the pre-classification pipeline and the post-classification pipeline. The pre-classification pipeline first performs classification on each video snippet and then aggregate the snippet-level classification scores to obtain the video-level classification score. In contrast, the post-classification pipeline aggregates the snippet-level features first and then predicts the video-level classification score based on the aggregated feature. Although the classifiers in these two pipelines are used in different ways, the role they play is exactly the same---to classify the given features to identify the corresponding action categories. To this end, an ideal classifier can make both pipelines work. This inspires us to simultaneously learn these two pipelines in a unified framework to obtain an effective classifier. Specifically, in the proposed learning framework, we implement two parallel network streams to model the two localization-by-classification pipelines simultaneously and make the two network streams share the same classifier. This achieves the novel Equivalent Classification Mapping (ECM) mechanism. Moreover, we discover that an ideal classifier may possess two characteristics: 1) The frame-level classification scores obtained from the pre-classification stream and the feature aggregation weights in the post-classification stream should be consistent; 2) The classification results of these two streams should be identical. Based on these two characteristics, we further introduce a weight-transition module and an equivalent training strategy into the proposed learning framework, which assists to thoroughly mine the equivalence mechanism. Comprehensive experiments are conducted on three benchmarks and ECM achieves accurate action localization results.
\end{abstract}

% Note that keywords are not normally used for peerreview papers.
}

% make the title area
\maketitle

\IEEEdisplaynontitleabstractindextext

\IEEEpeerreviewmaketitle

\IEEEraisesectionheading{\section{Introduction}\label{sec:introduction}}

\IEEEPARstart{T}{emporal} action localization aims to localize action instances from the given untrimmed video by determining the start temporal points, end temporal points, and the corresponding action categories. In the studied weakly supervised setting, the action localizers are learned directly from video-level labels, without requiring fine segment-level annotations.
\begin{figure}[htbp]
	\graphicspath{{figure/}}
	\centering
	\includegraphics[width=1\linewidth]{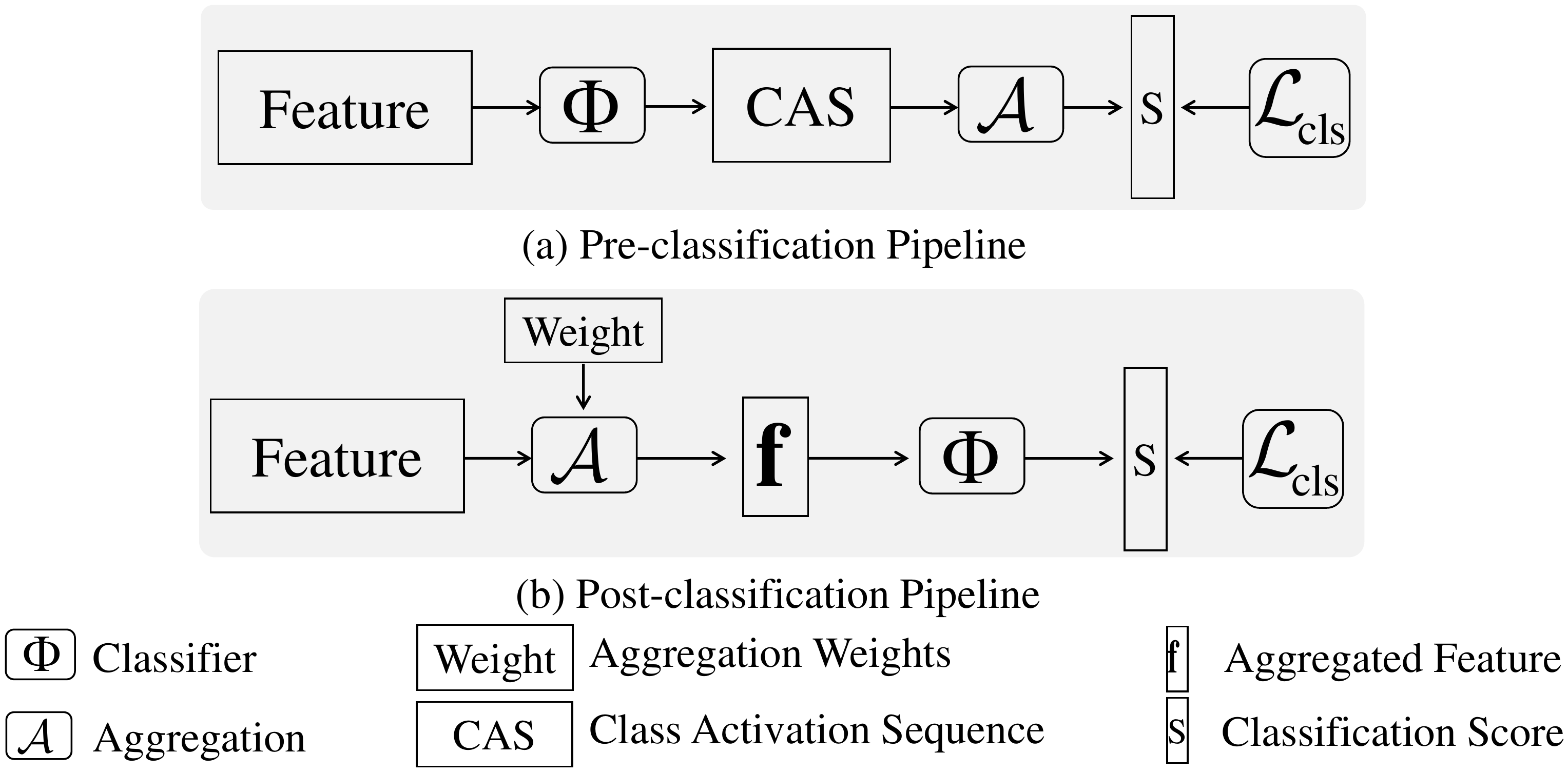}
	\caption{Illustration of two action localization pipelines. (a) Pre-classification pipeline first performs classification at each temporal point then aggregates scores. (b) Post-classification pipeline first aggregates features then predicts the classification score. Both pipelines are driven by the video-level classification loss $\mathcal{L}_{cls}$.}
	\label{framework}
\end{figure}
Consequently, weakly supervised temporal action localization can alleviate the burdensome and expensive human annotation. Moreover, it is potential to further realize the learning process on web-scale unlabeled videos and achieve breakthroughs.

In order to eliminate the ambiguity brought by the weak supervision, most previous works adopt the localization-by-classification pipelines to estimate the video-level classification score, which can be categorized into two main pipelines, i.e., the pre-classification pipeline (see Figure \ref{framework}(a)) and the post-classification pipeline (see Figure \ref{framework}(b)). Among the existing methods, UntrimmedNet \cite{wang2017untrimmednets} makes the pioneering exploration for the pre-classification pipeline. It first performs classification at each temporal point to obtain the \textit{class activation sequence}, which is then aggregated to predict the video-level classification score. Based on this work, CMCS \cite{liu2019completeness} proposes a diversity loss to model the completeness of actions. Meanwhile, 3C-Net \cite{narayan20193c} introduces action count cues to distinguish adjacent action sequences. In aforementioned pre-classification pipelines, the classification mapping functions are learned based on each temporal point and its corresponding local neighbors. Such a mechanism would be beneficial to capturing local contrast information but tends to be less effective in perceiving long-term connections within each given video.

Apart from pre-classification pipelines, there are also works, such as \cite{nguyen2018weakly, nguyen2019weakly, yuan2019marginalized}, which adopt the post-classification pipelines. These works first aggregate features from all temporal points to form the video-level feature representation, then they predict the video-level classification scores by classifying the aggregated feature representation. The advantage of these methods is that the aggregated feature can represent long-term relationships. However, when performing classification at each temporal point in the evaluation phase, the class activation sequences generated by these methods show insufficient discriminability to localize actions that belong to different categories.

From above discussions, we can observe that both the pre-classification pipeline and the post-classification pipeline aim at learning the effective classification mapping functions to predict the classification scores from the input features. The difference is that the pre-classification pipeline uses the classifier to perform classification on feature of each temporal point, while the post-classification pipeline uses the classifier to perform classification on the aggregated feature of the whole video sequence. Note that the post-classification pipeline uses \textit{weighted sum} to aggregate snippet features. By such a linear operation, the aggregated feature remains in the same feature space with point-level features. When an ideal classification mapping function is given, both pipelines would obtain accurate classification results. This inspires us to simultaneously learn these two pipelines under a newly proposed Equivalent Classification Mapping (ECM) mechanism to obtain the desired classifier. After this, the learned classifier can be used to identify the presence of actions on each temporal point.

The framework of ECM is shown in Figure \ref{classifier}. Compiling the basic equivalent classification mapping spirit, ECM adopts two parallel network streams to model the pre-classification pipeline and the post-classification pipeline, respectively, and makes these two network streams share the same classifier. This simple and direct implementation of the equivalent classification mapping mechanism brings adequate performance gains over both the pre-classification baseline and the post-classification baseline. However, there are still non-negligible performance gaps when compared with state-of-the-art methods, indicating that adequately mining the equivalence mechanism is critical for good results.

We propose two equivalence-based components to thoroughly explore the equivalence mechanism: the module and the equivalent training strategy. The weight-transition module is designed to learn precise attention weights for the post-classification stream. Specifically, ECM transits the frame-level classification scores obtained from the pre-classification stream to generate the feature aggregation weights for the post-classification pipeline. This is different from traditional post-classification pipelines \cite{nguyen2018weakly, nguyen2019weakly, yuan2019marginalized}, where the feature aggregation weights are inferred in self-attention-like manners. Moreover, apart from the basic classification losses of each network stream, we introduce a novel equivalent training strategy, which contains a classification-to-classification consistency loss and an aggregation-to-classification consistency loss. The former is designed to penalize the inconsistency of the classification scores from two network streams, while the latter is designed to penalize the inconsistency of the classification score and the aggregation weights. In summary, the weight-transition module and the equivalent training strategy assist to adequately mine the equivalence mechanism between the pre-classification stream and the post-classification stream. This improves the performance of ECM and results in accurate action localization results.

The contribution of this work can be summarized as follows:
\begin{itemize}
	\item ECM reveals the equivalence mechanism, indicating that both the classification for snippet features in the pre-classification pipeline and the classification for aggregated features in the post-classification pipeline pursue the same ideal classifier. Although conceptually simple, the equivalence mechanism is overlooked by previous methods, but plays an essential role in localizing actions under weak supervision.
	\item We propose two equivalence-based components, i.e., the weight-transition module and the equivalent training strategy, to adequately mine the equivalence mechanism. A simple and direct implementation of the equivalence mechanism shows improvements over baselines, which can be further promoted by these two equivalence-based components.
	\item Empirically, ECM starts from two simple and widely used baselines (with performance $19.6$ and $17.1$), employs equivalence-based components, explores the equivalence mechanism, and achieves accurate localization performance ($29.1$ on THUMOS14) without bells and whistles. Considering its simplicity, ECM can serve as a solid baseline for future studies.
\end{itemize}

%%%%%%%%% Related work
\section{Related Work}

\subsection{Action recognition} Action recognition is a fundamental task for video analysis and understanding. With the development of deep learning, many effective algorithms have emerged. An early classic work is \cite{simonyan2014two}, which adopts a two-stream ConvNet architecture, to incorporate single frame RGB image and multi-frame optical flow. Then, C3D \cite{tran2015learning} uses 3D ConvNets for spatiotemporal feature
learning. Next, I3D \cite{carreira2017quo} proposes a new Two-Stream Inflated 3D ConvNet and P3D ResNet \cite{qiu2017learning} fully exploits ResNet by simulating 3D CNN. Recently, Feichtenhofer \textit{et al.} propose SlowFast  \cite{feichtenhofer2019slowfast} networks for video recognition. However, all these methods learn from trimmed videos, but real applications usually encounter untrimmed videos.

\begin{figure*}[htbp]
	\graphicspath{{figure/}}
	\centering
	\includegraphics[width=0.80\linewidth]{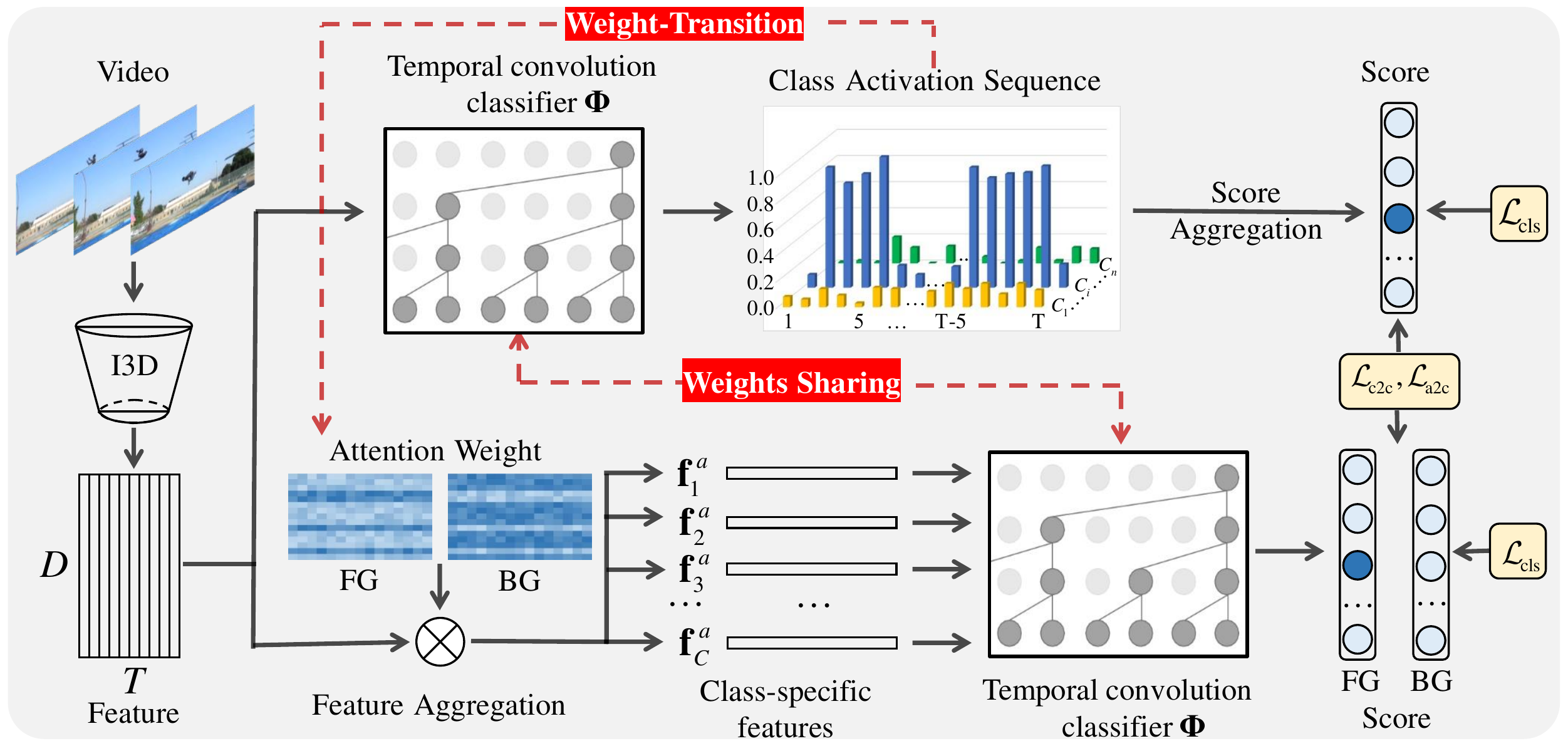}
	\caption{Framework of the proposed equivalent classification mapping (ECM) method. We extract video feature and send it to the pre-classification stream (shown on the top part) and the post-classification stream (shown on the bottom part). The pre-classification stream performs classification at each temporal point, while the post-classification stream performs classification at the aggregated video-level features. The classifier is shared between the pre-classification stream and the post-classification stream, and is learned by equivalence mechanism. The complete ECM is driven by two classification losses $\mathcal{L}_{\rm cls}$ and two equivalence-based losses ($\mathcal{L}_{\rm c2c}$ and $\mathcal{L}_{\rm a2c}$).}
	\label{classifier}
\end{figure*}

\subsection{Supervised action localization} Supervised action localization task learns from precisely annotated action instances and can deal with untrimmed videos which contain a large percentage of backgrounds. Early explorations adopt a detection-by-classification pipeline. S-CNN \cite{shou2016temporal} classifies sliding-window proposals to localize action instances. CDC \cite{shou2017cdc} and Lin \textit{et al.} \cite{lin2019bmn} predict the actionness score for each temporal point. Meanwhile, advances in object detection bring inspiration to action localization. Some methods follow Faster R-CNN \cite{ren2017faster} and perform two-stage action localization, e.g., R-C3D \cite{xu2017r} and TAL \cite{chao2018rethinking}. Similarly, some methods follow one-stage object detection methods \cite{liu2016ssd} and perform action localization, e.g., SSAD \cite{lin2017single} and GTAN \cite{long2019gaussian}. In addition, the recurrent memory module is used to capture long-term dependencies, such as SS-TAD \cite{buch2017end}, SST \cite{buch2017sst}. Apart from the above explorations, some noticeable works also include modeling temporal structure \cite{zhao2017temporal}, modeling context \cite{chao2018rethinking, dai2017temporal}, modeling relationships among action proposals \cite{zeng2019graph}, localizing action instances from a part of videos \cite{alwassel2018action}, etc. Recently, Yang \textit{et al.} propose A2Net \cite{yang2020revisiting} which adopts a novel anchor-free action localization module to tackle extremely short action instances or extremely long ones. G-TAD \cite{xu2020g} presents a graph convolutional network model to exploit video context and cast temporal action localization as a sub-graph detection problem. Zhao \textit{et al.} \cite{zhao2020bottom} introduce
two regularization terms to alleviate the problem of incorrect or
inconsistent predictions. In summary, supervised methods can explicitly learn from segment-level annotations and achieve accurate localizations. However, they are limited by the expensive annotations as well, which can be alleviated by the studied weakly supervised method.

\subsection{Weakly supervised action localization} Weakly supervised action localization task only requires video-level category information in the training phase. The previous methods can be divided into two categories, i.e., the pre-classification pipeline and the post-classification pipeline. In the pre-classification pipeline, UntrimmedNets \cite{wang2017untrimmednets} performs classification at each video snippet, then aggregates snippet scores at temporal dimension to obtain video-level classification scores. Later, W-TALC \cite{paul2018w} considers the co-activity similarity to model inter-video similarities and differences. Furthermore, CMCS \cite{liu2019completeness} proposes to jointly learn multiple classification networks and require them to generate diverse responses. Focused on the quality of the class activation sequence, there are some other promising works. CleanNet \cite{liu2019weakly} learns regression to adjust the action segments. TSM \cite{yu2019temporal} models the action structure via a multi-phase process. 3C-Net \cite{narayan20193c} introduces multi-label center loss to obtain discriminative feature representation. BaSNet \cite{lee2019background} tries to suppress the background response by giving different background labels on two branches. Recently, Gong \textit{et al.} \cite{gong2020learning} propose to learn class-specific and class-agnostic attention simultaneously. A2CL-PT \cite{min2020adversarial} adopts an adversarial approach to obtain more complete action instances. TSCN \cite{zhai2020two} presents two-stream consensus network to eliminate false positive action proposals and improve localization boundaries. EM-MIL \cite{luo2020weakly} explicitly models latent variables and adopts an expectation-maximization framework to better model the background information. In general, the foundation of pre-classification methods is the class activation sequence, generated by the classification mapping function.

Apart from the above methods, there is another pipeline, namely the post-classification pipeline. Inspired by the success of CAM \cite{zhou2016learning}, STPN \cite{nguyen2018weakly} proposes to first aggregate video features then perform classification. Later, Nguyen \textit{et al.} \cite{nguyen2019weakly} extend STPN \cite{nguyen2018weakly} by introducing background modeling and top-down class-guided attention. Meanwhile, MAAN \cite{yuan2019marginalized} develops the post-classification pipeline with a marginalized average aggregation module. Recently, Shi \textit{et al.} \cite{shi2020weakly} use conditional variational auto-encoder to model the frame-wise representation conditioned for tackling action-context confusion issue. In general, the post-classification pipeline is aware of the complete video features, but its response to each temporal point may be not discriminative enough.

%------------------------------------------------------------------------
\section{Method}

\begin{figure*}[htbp]
	\graphicspath{{figure/}}
	\centering
	\includegraphics[width=1\linewidth]{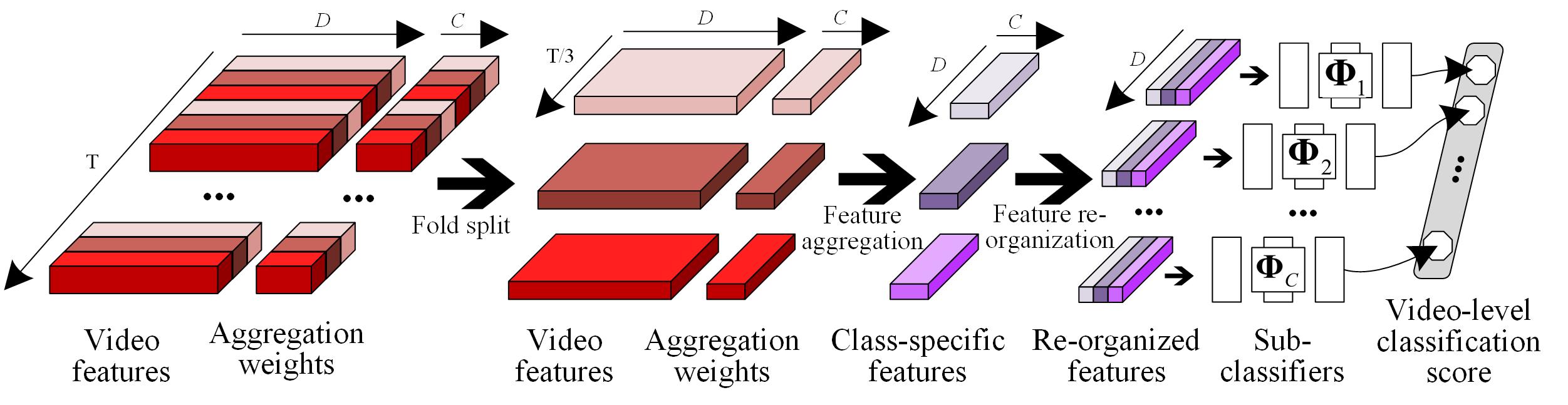}
	\caption{Illustration of the detailed operations for feature aggregation in the post-classification stream.}
	\label{aggregation}
\end{figure*}

Given a dataset containing $C$ action categories, each video in the dataset has the classification label ${\mathbf{y}}=[y_{1}, y_{2},...,y_{C}]$, where $y_{c} \in \{0, 1\}, c\in[1,C]$ indicates whether there is an action instance belonging to category $c$. Each video is firstly divided into $T$ snippets and then a feature extractor is used to extract the feature representation $\mathbf{f}_{t} \in \mathbb{R}^{D}, t\in[1,T]$ from each video snippet. Under the weakly supervised learning scenario, ECM learns the classifier $\bm{\Phi}=[\bm{\Phi}_{1}, \bm{\Phi}_{2}, ..., \bm{\Phi}_{C}]$ from the video-level classification label $\mathbf{y}$, $\bm{\Phi}_{c}$ is the specific sub-classifier for category $c$, aiming at discovering the temporal locations of the desired action instances for each video.

\subsection{The body network}
\label{chaper-method-network}

As shown in Figure \ref{classifier}, the body network consists of two streams: the pre-classification stream and the post-classification stream. They share the same classification mapping function but make predictions in different manners.

We start the elaboration of ECM from a concise mathematical proof that the point-level features in the pre-classification branch and aggregated video-level features in the post-classification branch lie in the same feature space, and they require an identical ideal classifier. Suppose the ideal classifier for pre-classification stream and post-classification stream is $\bm{\Phi}^{pre}$ and $\bm{\Phi}^{post}$, respectively. Given snippet features $\mathbf{F}=\{\mathbf{f}_{1},...,\mathbf{f}_{T}\}$, we adopt \textit{weighted sum} operation to aggregate features and obtain $\mathbf{f}^{agg}$, i.e., $\mathbf{f}^{agg}=\sum_{t=1}^{T} w_{t} \times \mathbf{f}_{t}$, where $w_{t}$ is the weight for $\mathbf{f}_{t}$. Because \textit{weighted sum} is a linear operation, $\mathbf{f}^{agg}$ is in the same feature space $\mathbb{F}$ with $\{\mathbf{f}_{1},...,\mathbf{f}_{T}\}$. Moreover, as for feature space $\mathbb{F}$, there exists one optimal classifier $\bm{\Phi}$ that can precisely classify most number of features. Because $\bm{\Phi}^{pre}=\bm{\Phi}=\bm{\Phi}^{post}$, the two streams aim to learn an identical ideal classifier.

To endow the classifier $\bm{\Phi}$ with the helpful temporal reception field, we use three temporal convolutional layers to build it (see Figure \ref{classifier}), where the kernel sizes of the first two layers are $3$ while it of the last layer is $1$. Here, the classifier is used to predict the classification scores for each temporal point, i.e., each video snippet. Specifically, in the pre-classification stream, original video features are directly passed through the classifier. Then, the network obtains classification scores along temporal points and generates class activation sequences for the whole video. After that, we adopt the top-$k$ mean strategy to aggregate the classification scores for each category $c$ and obtain the video-level classification score $\mathbf{s}^{e}=[s^{e}_{1}, s^{e}_{2}, ..., s^{e}_{C}]$.

In the post-classification stream, the inputs consist of both the video features $\mathbf{F}$ and the category-specific aggregation weights. Here, we explore both action weights ${\mathbf{W}^{a}} \in \mathbb{R}^{C \times T}$ and background weights ${\mathbf{W}^{b}} \in \mathbb{R}^{C \times T}$ to generate the category-specific action features $\mathbf{F}^{a}=[\mathbf{f}^{a}_{1}, \mathbf{f}^{a}_{2}, ..., \mathbf{f}^{a}_{C}]$ and category-specific background features $\mathbf{F}^{b}=[\mathbf{f}^{b}_{1}, \mathbf{f}^{b}_{2}, ..., \mathbf{f}^{b}_{C}]$, where $\mathbf{f}^{*}_{c} \in \mathbb{R}^{D \times 3}$.

In detail, the category-specific feature for either action or background is obtained by \textit{weighted sum} operation over the video features $\mathbf{F}=\{\mathbf{f}_{1},...,\mathbf{f}_{T}\}$:
\begin{equation}
\mathbf{f}^{a}_{c}=\sum_{t=1}^{T} w^{a}_{c,t} \times \mathbf{f}_{t}
\end{equation}
\begin{equation} \mathbf{f}^{b}_{c}=\sum_{t=1}^{T} w^{b}_{c,t} \times \mathbf{f}_{t}
\end{equation}
where $w^{a}_{c,t}$ and $w^{b}_{c,t}$ are elements in $\mathbf{W}^{a}$ and $\mathbf{W}^{b}$, respectively. Then, each sub-classifier $\bm{\Phi}_{c}$ is applied to the corresponding category-specific action feature $\mathbf{f}^{a}_{c}$ to obtain the final classification score $\mathbf{s}^{o}=[s^{o}_{1}, s^{o}_{2}, ..., s^{o}_{C}]$, where $s^{o}_{c}=f(\mathbf{f}^{a}_{c}|\bm{\Phi}_{c})$, $f(\cdot)$ denotes the network forward operation.

The classifier is shared between the pre-classification stream and the post-classification stream, where the first layer adopts temporal convolution with kernel size $3$. Consequently, the post-classification stream requires the temporal length of the aggregated feature to be $3$. Given video features $\mathbf{F}=\{\mathbf{f}_{1},...,\mathbf{f}_{T}\}$ and class-specific aggregation weights, a direct \textit{weighted sum} would generate one aggregated feature. In order to aggregate $3$ features for each category, we propose a specific feature aggregation strategy, as shown in Figure \ref{aggregation}. First of all, we evenly divide video features as well as the corresponding aggregation weights into three folds by sampling every three snippets. Then, we perform feature aggregation and generate $3$ features for each category. After that, feature vectors belonging to the same category are re-organized together. Finally, category-specific features are sent to the corresponding sub-classifier $\bm{\Phi}_{c}$ to predict video-level classification scores. Comparing to the strategy that directly extracting the whole video features in one fold, the aggregated video-level features extracted by our strategy can better fit the input structure of the classifier, thus facilitating a more effective equivalent learning scheme.

\begin{figure*}[htbp]
	\graphicspath{{figure/}}
	\centering
	\includegraphics[width=1\linewidth]{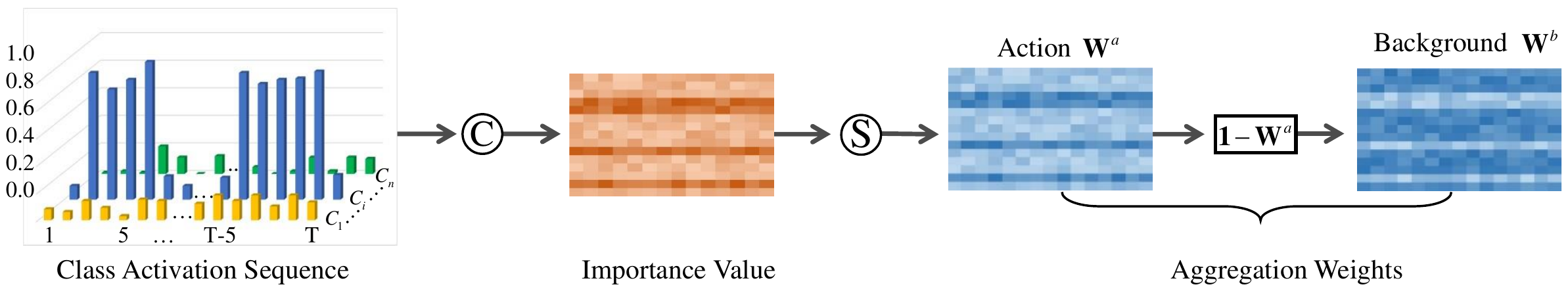}
	\caption{Equivalent weight-transition module. We predict the foreground and background aggregation weights from the class activation sequence. $\textcircled{c}$ indicates temporal convolution, $\circledS$ indicates sigmoid activation.}
	\label{attention}
\end{figure*}

\subsection{Equivalent weight-transition module}
The post-classification stream requires category-specific aggregation weights $\mathbf{W}^{a}$ and $\mathbf{W}^{b}$ to aggregate features. Considering that the class activation sequence obtained from the pre-classification stream can reveal the probability that a temporal point belongs to an action instance, we propose to obtain the aggregation weights by transiting the class activation sequence, as shown in Figure \ref{attention}. We use a convolutional layer with kernel size $1$ to predict the importance values for each temporal point. After that, we apply the sigmoid activation and obtain the action aggregation weights $\mathbf{W}^{a}$, while the background aggregation weights can be obtained via $\mathbf{W}^{b}=\mathbf{1}-\mathbf{W}^{a}$. Compared with the conventional methods \cite{nguyen2018weakly, nguyen2019weakly, yuan2019marginalized, lee2019background}, the proposed equivalent weight-transition module shows two characteristics. Firstly, existing works adopt an extra network to predict the aggregation weights, while the proposed module can directly obtain aggregation weights from the pre-classification stream. Besides, existing methods predict category-agnostic weights and use them to obtain one aggregated feature to represent the input video. In contrast, we learn category-specific weights and use them to obtain aggregated features for each different action categories. Such category-specific features would bring richer representation and assist ECM to clearly distinguish different action categories, as verified in Section \ref{chapter-ablation-exp}.

\subsection{Classification-to-classification consistency}
The equivalent training strategy consists of a classification-to-classification consistency loss $\mathcal{L}_{\rm c2c}$ and an aggregation-to-classification consistency loss $\mathcal{L}_{\rm a2c}$. Given an input video, the pre-classification stream predicts the classification score $\mathbf{s}^{e}$, while the post-classification stream predicts the classification score $\mathbf{s}^{o}$ with action aggregation weights $\mathbf{W}^{a}$. Based on the intuition that the classification scores predicted by the two network streams should be identical for the same input video, we introduce the classification-to-classification consistency loss $\mathcal{L}_{\rm c2c}$ to the learning process:
\begin{equation}
\mathcal{L}_{\text {c2c}}=\frac{1}{C} \sum_{i=1}^{C} (s^{\rm e}_{i} - s^{\rm o}_{i})^{2}
\end{equation}

The classification-to-classification consistency term $\mathcal{L}_{\text {c2c}}$ plays a role in making one stream perceive the predictions from the other stream. Despite the simplicity, it obviously facilitates the learning of the classification mapping function, as shown in Section \ref{chapter-ablation-exp}.
% i.e., $\mathcal{L}_{\rm equ}=\mathcal{L}_{\rm c2c} + \alpha \mathcal{L}_{\rm a2c}$, where $\alpha$ is the coefficient.

\subsection{Aggregation-to-classification consistency}
Besides classification-to-classification consistency, we further explore the aggregation-to-classification consistency loss. The motivation is that the video-level classification score should be consistent with the aggregation weights. Specifically, the proposed aggregation-to-classification loss $\mathcal{L}_{\rm a2c}$ not only requires the action attention weights $\mathbf{W}^{a}$ to highlight snippets within an action, but also requires the background attentions  $\mathbf{W}^{b}=\mathbf{1}-\mathbf{W}^{a}$ to exclude all action snippets. It assists the learning of the attention weights with considerations about both action presence and action absence. This training strategy is different from conventional classification loss, where the attention weights may only highlight the most discriminative snippets within an action but still can correctly predict video-level classification label.

Considering a video with classification label $\mathbf{y}$, the post-classification stream can predict the action presence score $\mathbf{s}^{o}=[s^{o}_{1}, s^{o}_{2}, ..., s^{o}_{C}]$ with action weights $\mathbf{W}^{a}$. Meanwhile, it can predict the action absence score $ \tilde{\mathbf{s}}^{o}=[\tilde{s}^{o}_{1}, \tilde{s}^{o}_{2}, ..., \tilde{s}^{o}_{C}]$ with background weights $\mathbf{W}^{b}$. Similar to the action presence score $\mathbf{s}^{o}$, the action absence score $\tilde{\mathbf{s}}^{o}$ is obtained by $\tilde{\mathbf{s}}^{o}_{c}=f(\mathbf{f}^{b}_{c}|\bm{\Phi}_{c})$. Then, for an input video that contains $k$ action categories, we select classification scores for presenting action categories from $\mathbf{s}^{o}$ and $\tilde{\mathbf{s}}^{o}$ according to classification label $\mathbf{y}$. After that, we define $\mathbf{s}'=[s^{o}_{k_{1}}, s^{o}_{k_{2}}, ..., s^{o}_{k_{k}}, \tilde{s}^{o}_{k_{1}}, \tilde{s}^{o}_{k_{2}}, ..., \tilde{s}^{o}_{k_{k}}]$, which corresponds to $k$ positive labels and $k$ negative labels. Finally, the aggregation-to-classification consistency loss $\mathcal{L}_{\rm a2c}$ can be calculated as follows:
\begin{equation}
\mathcal{L}_{\rm a2c}=-\frac{1}{2k}[\sum_{i=1}^{k} {\rm log}(s^{o}_{k_{i}}) + \sum_{i=1}^{k} {\rm log}(1-\tilde{s}^{o}_{k_{i}}))]
\end{equation}

The proposed aggregation-to-classification consistency training strategy can guide ECM to better distinguish action segments and backgrounds. Besides, the existing background modeling strategy \cite{nguyen2019weakly} and background suppression strategy \cite{lee2019background} can be regarded as the specific cases of this strategy, where they only consider the background category when adjusting attention weights.

\subsection{Training and inference}

% Table generated by Excel2LaTeX from sheet 'anet1.2-complete'
\begin{table*}[htbp]
	\centering
	\caption{Comparisons between ECM and state-of-the-art methods on ActivityNet v1.2 dataset. The performance of both fully supervised and weakly supervised methods are reported. We report mAP under different thresholds, as well as the average mAP.}
	\setlength{\tabcolsep}{3pt}
	\begin{tabular}{c|c|cc|c|cccccccccc|c}
		\toprule
		\multicolumn{2}{c|}{Sup.} & \multicolumn{1}{c|}{Method} & Pub.  & feature & 0.50  & 0.55  & 0.60  & 0.65  & 0.70  & 0.75  & 0.80  & 0.85  & 0.90  & 0.95  & avg (0.50:0.95) \\
		\midrule
		\multicolumn{2}{c|}{Full} & \multicolumn{1}{c|}{SSN \cite{zhao2017temporal}} & ICCV 17 & TS    & 41.3  & -     & -     & -     & -     & 27.0  & -     & -     & -     & 6.1   & \textbf{26.6} \\
		\midrule
		\multirow{12}[6]{*}{Weak} & \multicolumn{1}{c|}{\multirow{9}[2]{*}{Pre-Cls}} & \multicolumn{1}{c|}{AutoLoc \cite{shou2018autoloc}     } & ECCV 18 & UNT   & 27.3  & 24.9  & 22.5  & 19.9  & 17.5  & 15.1  & 13.0  & 10.0  & 6.8   & 3.3   & 16.0 \\
		&       & \multicolumn{1}{c|}{TSM \cite{yu2019temporal}      } & ICCV 19 & I3D   & 28.3  & 26.0  & 23.6  & 21.2  & 18.9  & 17.0  & 14.0  & 11.1  & 7.5   & 3.5   & 17.1 \\
		&       & \multicolumn{1}{c|}{W-TALC \cite{paul2018w}           } & ECCV 18 & I3D   & 37.0  & -     & -     & -     & 14.6  & -     & -     & -     & -     & -     & 18.0 \\
		&       & \multicolumn{1}{c|}{EM-MIL \cite{luo2020weakly}} & ECCV 20 & I3D   & 37.4  & -     & -     & -     & 23.1  & -     & -     & -     & 2.0   & -     & 20.3 \\
		&       & \multicolumn{1}{c|}{CleanNet \cite{liu2019weakly}       } & ICCV 19 & I3D   & 37.1  & 33.4  & 29.9  & 26.7  & 23.4  & 20.3  & 17.2  & 13.9  & 9.2   & 5.0   & 21.6 \\
		&       & \multicolumn{1}{c|}{3C-Net \cite{narayan20193c}       } & ICCV 19 & I3D   & 37.2  & -     & -     & -     & 23.7  & -     & -     & -     & 9.2   & -     & 21.7 \\
		&       & \multicolumn{1}{c|}{CMCS \cite{liu2019completeness} } & CVPR 19 & I3D   & 36.8  & -     & -     & -     & -     & 22.0  & -     & -     & -     & 5.6   & 22.4 \\
		&       & \multicolumn{1}{c|}{BaSNet \cite{lee2019background}   } & AAAI 20 & I3D   & 38.5  & -     & -     & -     & -     & 24.2  & -     & -     & -     & 5.6   & 24.3 \\
		&       & \multicolumn{1}{c|}{ACL \cite{gong2020learning}} & CVPR 20 & I3D   & 40.0  & -     & -     & -     & -     & 25.0  & -     & -     & -     & 4.6   & 24.6 \\
		\cmidrule{2-16}          & \multirow{2}[2]{*}{Post-Cls} & \multicolumn{1}{c|}{DGAM \cite{shi2020weakly}} & CVPR 20 & I3D   & \textbf{41.0} & 37.5  & 33.5  & 30.1  & 26.9  & 23.5  & 19.8  & 15.5  & 10.8  & 5.3   & 22.4 \\
		&       & \multicolumn{1}{c|}{TSCN \cite{zhai2020two}} & ECCV 20 & I3D   & 37.6  & -     & -     & -     & -     & 23.7  & -     & -     & -     & 5.7   & 23.6 \\
		\cmidrule{2-16}          & Equ.  & \multicolumn{2}{c|}{ECM} & I3D   & \textbf{41.0} & \textbf{37.7} & \textbf{34.2} & \textbf{31.5} & \textbf{28.5} & \textbf{24.9} & \textbf{21.2} & \textbf{17.0} & \textbf{12.1} & \textbf{6.5} & \textbf{25.5} \\
		\bottomrule
	\end{tabular}%
	\label{tab:cmp-anet1.2}
\end{table*}%

In training, we calculate the classification loss $\mathcal{L}_{\rm cls,e}$ and $\mathcal{L}_{\rm cls,o}$ for the pre-classification stream and the post-classification stream, respectively. Formally, the classification problem can be formulated as a multi-label classification problem. We perform $L1-$normalization on the original classification label $\mathbf{y}$ and calculate the cross-entropy loss. Besides, the classification-to-classification consistency loss $\mathcal{L}_{\rm c2c}$ and the aggregation-to-classification consistency loss $\mathcal{L}_{\rm a2c}$ are also calculated to guide the mining of the equivalence mechanism. Thus, the complete loss function can be calculated as:
\begin{equation}
\mathcal{L}=\mathcal{L}_{\rm cls,e} + \mathcal{L}_{\rm cls,o} + \alpha \mathcal{L}_{\rm c2c} + \beta \mathcal{L}_{\rm a2c}
\label{total_loss}
\end{equation}
where $\alpha$ and $\beta$ are trade-off coefficients to balance the effect of different losses.

When the training process is complete, we forward each input video through the learned classifier and obtain the class activation sequences. Then, following BaSNet \cite{lee2019background}, we discard categories whose classification scores are smaller than a threshold $\tau$. After that, the remaining class activation sequences are min-max normalized along the temporal dimension, which is followed by the temporal actionness grouping operation \cite{zhao2017temporal} to localize action instances. Finally, redundant action instances are filtered out via NMS.

\section{Experiments}

\label{section-experiments}

\subsection{Experimental setups}
\textbf{Dataset.} We perform experiments on three benchmarks: THUMOS14 \cite{THUMOS14}, ActivityNet v1.2 \cite{caba2015activitynet} and ActivityNet v1.3 \cite{caba2015activitynet}. THUMOS14 consists of $20$ action categories, including $200$ videos for training and $213$ videos for testing. ActivityNet v1.2 consists of $100$ action categories, $9682$ videos. The video number ratio among training, validation and testing sets is $2$:$1$:$1$. ActivityNet v1.3 is an extension of ActivityNet v1.2, with $200$ categories and $19994$ videos.

\textbf{Metric.} The evaluation is performed under the official metric of each dataset, i.e., mean Average Precision (mAP). THUMOS14 focuses on mAP under threshold $0.5$, while ActivityNet focuses on the average mAP under thresholds [$0.50$:$0.05$:$0.95$]. Following previous works \cite{nguyen2018weakly, nguyen2019weakly, lee2019background}, we adopt official evaluation tools of the ActivityNet dataset to perform evaluations.

\textbf{Feature extraction.}
To extract features, each video is evenly divided into $T$ snippets and we uniformly sample $16$ frames from each snippet, similar to previous works \cite{nguyen2018weakly, nguyen2019weakly}. The optical flow is calculated via the TV-L1 \cite{zach2007duality} algorithm. I3D model \cite{carreira2017quo} pre-trained on the Kinetics-400 dataset is used to extract video features, without finetuning on THUMOS14 \cite{THUMOS14} or ActivityNet v1.3 \cite{caba2015activitynet}. We extract both the appearance features and the motion features, which are concatenated together to represent the video sequence. The concatenated feature dimension is $2048$.  Finally, the snippet number for THUMOS14, ActivityNet v1.2 and ActivityNet v1.3 are $750$, $100$ and $100$, respectively.

\textbf{Training and evaluation details.}
ECM is implemented using PyTorch \cite{paszke2019pytorch}. Adam \cite{kingma2014adam} solver is used to optimize the network. For all experiments, we set batch size to $16$, the learning rate to $2 \times 10^{-4}$ and the weight decay to $5 \times 10^{-4}$. We train ECM with $150$ epochs, $40$ epochs and $40$ epochs for THUMOS14, ActivityNet v1.2 and ActivityNet v1.3, respectively. The parameters are empirically determined via gird search. Specifically, we set $k=1/8$ for the top-$k$ mean aggregation. The balance coefficients are $\alpha=0.05$ and $\beta=5$. In evaluation, the threshold to reject absent categories is $\tau=0.25$.

\subsection{Comparison with state-of-the-arts}

\textbf{Experiments on ActivityNet v1.2.} Table \ref{tab:cmp-anet1.2} reports the performance on ActivityNet v1.2 dataset. Most methods adopt the I3D feature, while there are also UntrimmedNet features (UNT) \cite{wang2017untrimmednets}. The weakly supervised WTAL is developed by both the pre-classification pipeline and the post-classification pipeline. AutoLoc \cite{shou2018autoloc} firstly builds the foundation performance of $16.0$. At the same time, W-TALC \cite{paul2018w} achieves $18.0$ via exploring co-activity relationship. Later, CleanNet \cite{liu2019weakly} proposes to learn regression. 3C-Net \cite{narayan20193c} strives to lean discriminative features via constraining centerness and action counts. CMCS \cite{liu2019completeness} simultaneously learns multiple complementary class activation sequences. These above methods gradually improve the performance to $22.4$ \cite{liu2019completeness}.

\begin{table}[htbp]
	\centering
	\caption{Comparison experiments on ActivityNet v1.3 dataset. We report mAP under thresholds $\{0.50$, $0.75$, $0.95\}$, as well as the average mAP. "pre" indicates pre-classification methods, "post" indicates post-classification method.}
	\setlength{\tabcolsep}{2.5pt}
	\begin{tabular}{c|ccc|c|cccc}
		\toprule
		\multicolumn{2}{c|}{Sup.} & \multicolumn{1}{c|}{Method} & Pub.  & Fea.  & 0.5   & 0.75  & 0.95  & 0.50:0.95 \\
		\midrule
		\multicolumn{2}{c|}{\multirow{8}[2]{*}{Full}} & \multicolumn{1}{c|}{TCN \cite{dai2017temporal}     } & ICCV 17 & TS    & 36.4  & 21.2  & 3.9   & - \\
		\multicolumn{2}{c|}{} & \multicolumn{1}{c|}{TAL \cite{chao2018rethinking}  } & CVPR 18 & I3D   & 38.2  & 18.3  & 1.3   & 20.2  \\
		\multicolumn{2}{c|}{} & \multicolumn{1}{c|}{CDC \cite{shou2017cdc}         } & CVPR 17 & -     & 45.3  & 26.0  & 0.2   & 23.8  \\
		\multicolumn{2}{c|}{} & \multicolumn{1}{c|}{TSA-Net \cite{zhao2020bottom}} & ECCV 20 & I3D   & 43.5  & 33.9  & \textbf{9.2} & 30.1  \\
		\multicolumn{2}{c|}{} & \multicolumn{1}{c|}{PGCN \cite{zeng2019graph}       } & ICCV 19 & I3D   & 48.3  & 33.2  & 3.3   & 31.1  \\
		\multicolumn{2}{c|}{} & \multicolumn{1}{c|}{BMN \cite{lin2019bmn}          } & ICCV 19 & TS    & 50.1  & 34.8  & 8.3   & 33.9  \\
		\multicolumn{2}{c|}{} & \multicolumn{1}{c|}{G-TAD \cite{xu2020g}} & CVPR 20 & I3D   & 50.4  & \textbf{34.6} & 9.0   & 34.1  \\
		\multicolumn{2}{c|}{} & \multicolumn{1}{c|}{GTAN \cite{long2019gaussian}    } & CVPR 19 & P3D   & \textbf{52.6} & 34.1  & 8.9   & \textbf{34.3} \\
		\midrule
		\multirow{10}[6]{*}{Weak} & \multicolumn{1}{c|}{\multirow{4}[2]{*}{pre}} & \multicolumn{1}{c|}{TSM \cite{yu2019temporal}         } & ICCV 19 & I3D   & 30.0  & 19.0  & 4.5   & - \\
		& \multicolumn{1}{c|}{} & \multicolumn{1}{c|}{CMCS \cite{liu2019completeness}    } & CVPR 19 & I3D   & 34.0  & 20.9  & 5.7   & 21.2  \\
		& \multicolumn{1}{c|}{} & \multicolumn{1}{c|}{BaSNet \cite{lee2019background}      } & AAAI 20 & I3D   & 34.5  & 22.5  & 4.9   & 22.2  \\
		& \multicolumn{1}{c|}{} & \multicolumn{1}{c|}{ACL \cite{gong2020learning}} & ECCV 20 & I3D   & \textbf{36.8} & 22.0  & 5.2   & 22.5  \\
		\cmidrule{2-9}          & \multicolumn{1}{c|}{\multirow{5}[2]{*}{post}} & \multicolumn{1}{c|}{STPN \cite{nguyen2018weakly}       } & CVPR 18 & I3D   & 29.3  & 16.9  & 2.6   & - \\
		& \multicolumn{1}{c|}{} & \multicolumn{1}{c|}{LTSR \cite{zhang2019learning}      } & AAAI 19 & Res   & 33.1  & 18.7  & 3.3   & 21.8  \\
		& \multicolumn{1}{c|}{} & \multicolumn{1}{c|}{MAAN \cite{yuan2019marginalized}   } & ICLR 19 & I3D   & 33.7  & 21.9  & 5.5   & - \\
		& \multicolumn{1}{c|}{} & \multicolumn{1}{c|}{TSCN \cite{zhai2020two}} & ECCV 20 & I3D   & 35.3  & 21.4  & 5.3   & 21.7  \\
		& \multicolumn{1}{c|}{} & \multicolumn{1}{c|}{WSBM \cite{nguyen2019weakly}       } & ICCV 19 & I3D   & 36.4  & 19.2  & 2.9   & - \\
		\cmidrule{2-9}          & \multicolumn{3}{c|}{ECM} & I3D   & 36.7  & \textbf{23.6} & \textbf{5.9} & \textbf{23.5} \\
		\bottomrule
	\end{tabular}%
	\label{tab:cmp-anet1.3}
\end{table}%

Recently, BaSNet \cite{lee2019background} proposes to suppress the response of backgrounds. They simultaneously learns two streams, i.e., the base stream and the suppression stream, which share network weights but dispose of different input data and pursue different classification target for the background category. Although BaSNet \cite{lee2019background} maintains two streams and share weights between them, both two streams essentially adopt the pre-classification pipeline. Experimentally, there are $1.2$ performance gaps between BaSNet \cite{lee2019background} and ECM. This demonstrates the equivalence mechanism is more applicable than the individual pre-classification mechanism for learning a powerful classifier. Beyond, ACL \cite{gong2020learning} achieves $24.6$ and shows high performance. Based on the pre-classification pipeline, ACL \cite{gong2020learning} aggregates features from all temporal points, constructs triplet features and calculates cluster-based loss, so as to enhance discriminability. However, ACL \cite{gong2020learning} only relies on the pre-classification pipeline to learn the classifier and ignores the post-classification pipeline. In contrast, ECM not only aggregates features to form a global representation, but also reveals the equivalence mechanism. Learning the same classifier from both pre-classification pipeline and post-classification pipeline, ECM exceeds ACL \cite{gong2020learning} and builds new state-of-the-art performance.

Apart from pre-classification pipeline, DGAM \cite{shi2020weakly} and TSCN \cite{zhai2020two} are two recently developed post-classification methods. DGAM \cite{shi2020weakly} starts from WSBM \cite{nguyen2019weakly} and optimizes attention weights via alternatively learning the generative attention module and the discriminative attention module. TSCN \cite{zhai2020two} adopts the pseudo ground truth to guide the learning of the attention weights and perform iterative refinement. Nevertheless, both DGAM \cite{shi2020weakly} and TSCN \cite{zhai2020two} only explore the post-classification pipeline to learn the classifier. In contrast, ECM demonstrates obvious superiority by exploring the equivalence mechanism between the pre-classification pipeline and the post-classification pipeline.

In addition to weakly supervised methods, ActivityNet is also explored by fully supervised method, e.g., structured segment network \cite{zhao2017temporal}. SSN \cite{zhao2017temporal} proposes a pyramid network to model temporal structures of action instances. It is encouraging that ECM only mines video-level classification labels but achieves competitive performance with SSN \cite{zhao2017temporal}. This further demonstrates the efficiency of the proposed equivalence classification mapping mechanism.

\textbf{Experiments on ActivityNet v1.3.} The comparison experiments on ActivityNet v1.3 dataset is shown in Table \ref{tab:cmp-anet1.3}. Temporal action localization on ActivityNet v1.3 is thoroughly studied in recent years, where there are multiple fully supervised and weakly supervised methods. Pioneering fully supervised methods \cite{dai2017temporal, shou2017cdc} adopt the detection-by-classification strategy and achieve average mAP $23.8$ \cite{shou2017cdc}. Later, two-stage methods \cite{chao2018rethinking, zhao2020bottom, lin2019bmn} improve the performance to $33.9$ step-by-step. Besides, GTAN \cite{long2019gaussian} is a representative one-stage method with high performance of $34.3$. Recently, the graph-convolution method \cite{zeng2019graph, xu2020g} exhibits superiority on temporal action localization, and achieves the performance of $34.1$ \cite{xu2020g}.

As for weakly supervised methods, for one thing, BaSNet \cite{lee2019background} and ACL \cite{gong2020learning} are representative pre-classification methods, achieving  the performance of $22.2$ and $22.5$, respectively. For another thing, STPN \cite{nguyen2018weakly} makes an early exploration and proposes the post-classification pipeline. This is further developed by subsequent works, e.g., LTSR \cite{zhang2019learning} achieves the performance of $21.8$ and TSCN \cite{zhai2020two} achieves the performance of $21.7$.

In contrast to existing weakly supervised methods, ECM achieves average mAP $23.5$ and exhibits obvious performance gains over existing strong competitors. In detail, ECM reaches the performance of $23.6$ and $5.9$ under threshold $0.75$ and $0.95$, respectively, showing high performance. Compared with fully supervised methods, on the one hand, ECM only learns from video-level classification labels but exceeds TCN \cite{dai2017temporal} and TAL \cite{chao2018rethinking} which learn from instance-level annotations. On the other hand, there is obvious performance gap between ECM and state-of-the-art supervised methods (e.g., G-TAD \cite{xu2020g}, GTAN \cite{long2019gaussian}), indicating the weakly supervised algorithm should be continuously developed.

\begin{table*}[htbp]
	\centering
	\caption{Comparisons between ECM and recent state-of-the-art methods on THUMOS14 dataset. We report mAP under different thresholds, where the comparison focuses on mAP@$0.5$.}
	\begin{tabular}{c|c|cc|c|ccccccc}
		\toprule
		Sup.  &       & \multicolumn{1}{c|}{Method} & Pub.  & Fea.  & 0.1   & 0.2   & 0.3   & 0.4   & 0.5   & 0.6   & 0.7  \\
		\midrule
		\multirow{7}[6]{*}{Full} & \multirow{2}[2]{*}{One Stage} & \multicolumn{1}{c|}{CDC \cite{shou2017cdc}          } & CVPR 17 & -     & -     & -     & 40.1  & 29.4  & 23.3  & 13.1  & 7.9 \\
		&       & \multicolumn{1}{c|}{GTAN \cite{long2019gaussian}     } & CVPR 19 & P3D   & 69.1  & 63.7  & 57.8  & 47.2  & 38.8  & -     & - \\
		\cmidrule{2-12}          & \multirow{3}[2]{*}{Two Stage} & \multicolumn{1}{c|}{R-C3D \cite{xu2017r}              } & ICCV 17 & -     & 54.5  & 51.5  & 44.8  & 35.6  & 28.9  & -     & - \\
		&       & \multicolumn{1}{c|}{TSA-Net \cite{zhao2020bottom}} & ECCV 20 & I3D   & -     & -     & 53.9  & 50.7  & 45.4  & 38.0  & 28.5 \\
		&       & \multicolumn{1}{c|}{TAL \cite{chao2018rethinking}   } & CVPR 18 & I3D   & 59.8  & 57.1  & 53.2  & 48.5  & 42.8  & 33.8  & 20.8 \\
		\cmidrule{2-12}          & \multirow{2}[2]{*}{Graph Based} & \multicolumn{1}{c|}{G-TAD \cite{xu2020g}} & CVPR 20 & TS    & -     & -     & 54.5  & 47.6  & 40.2  & 30.8  & 23.4 \\
		&       & \multicolumn{1}{c|}{PGCN \cite{zeng2019graph}        } & ICCV 19 & I3D   & 69.5  & 67.8  & 63.6  & 57.8  & $\textbf{49.1}$ & -     & - \\
		\midrule
		\multirow{13}[6]{*}{Weak} & \multirow{7}[2]{*}{Pre-Classification} & \multicolumn{1}{c|}{UntrimmedNet \cite{wang2017untrimmednets} } & CVPR 17 & UNT   & 44.4  & 37.7  & 28.2  & 21.1  & 13.7  & -     & - \\
		&       & \multicolumn{1}{c|}{AutoLoc \cite{shou2018autoloc}        } & ECCV 18 & UNT   & -     & -     & 35.8  & 29.0  & 21.2  & 13.4  & 5.8 \\
		&       & \multicolumn{1}{c|}{W-TALC \cite{paul2018w}              } & ECCV 18 & I3D   & 55.2  & 49.6  & 40.1  & 31.1  & 22.8  & -     & - \\
		&       & \multicolumn{1}{c|}{CMCS \cite{liu2019completeness}    } & CVPR 19 & I3D   & 57.4  & 50.8  & 41.2  & 32.1  & 23.1  & 15.0  & 7.0 \\
		&       & \multicolumn{1}{c|}{CleanNet \cite{liu2019weakly}          } & ICCV 19 & I3D   & -     & -     & 37.0  & 30.9  & 23.9  & 13.9  & 7.1 \\
		&       & \multicolumn{1}{c|}{TSM \cite{yu2019temporal}         } & ICCV 19 & I3D   & -     & -     & 39.5  & 31.9  & 24.5  & 13.8  & 7.1 \\
		&       & \multicolumn{1}{c|}{BaSNet \cite{lee2019background}      } & AAAI 19 & I3D   & 58.2  & 52.3  & 44.6  & 36.0  & 27.0  & 18.6  & 10.4 \\
		\cmidrule{2-12}          & \multirow{5}[2]{*}{Post-Classification} & \multicolumn{1}{c|}{STPN \cite{nguyen2018weakly}       } & CVPR 18 & I3D   & 52.0  & 44.7  & 35.5  & 25.8  & 16.9  & 9.9   & 4.3 \\
		&       & \multicolumn{1}{c|}{MAAN \cite{yuan2019marginalized}   } & ICLR 19 & I3D   & 59.8  & 50.8  & 41.1  & 30.6  & 20.3  & 12.0  & 6.9 \\
		&       & \multicolumn{1}{c|}{WSBM \cite{nguyen2019weakly}       } & ICCV 19 & I3D   & 64.2  & 59.5  & 49.1  & 38.4  & 27.5  & 17.3  & 8.6 \\
		&       & \multicolumn{1}{c|}{TSCN \cite{zhai2020two}} & ECCV 20 & I3D   & 63.4  & 57.6  & 47.8  & 37.7  & 28.7  & 19.4  & 10.2 \\
		&       & \multicolumn{1}{c|}{DGAM \cite{shi2020weakly}} & CVPR 20 & I3D   & 60.0  & 54.2  & 46.8  & 38.2  & 28.8  & 19.8  & 11.4 \\
		\cmidrule{2-12}          & Equivalent & \multicolumn{2}{c|}{ECM} & I3D   & 62.6  & 55.1  & 46.5  & 38.2  & $\textbf{29.1}$ & 19.5  & 10.9 \\
		\bottomrule
	\end{tabular}%
	\label{tab:cmp-thumos}%
\end{table*}%

\textbf{Experiments on THUMOS14.} In Table \ref{tab:cmp-thumos}, we compare ECM with recent state-of-the-art temporal action localization methods, including both the weakly supervised methods and the supervised ones on THUMOS14 dataset.

The weakly supervised methods consist of pre-classification methods and post-classification methods, both of which are widely explored. The first baseline of pre-classification methods is built by UntrimmedNet \cite{wang2017untrimmednets}. It learns video recognition model from video-level classification labels of untrimmed videos, which additionally discovers action instances via thresholding class activation sequences and achieves mAP@$0.5$=$13.7$. After this, AutoLoc \cite{shou2018autoloc} proposes the outer-inner-contrastive loss to precisely determine boundaries for actions and reaches $21.2$. The outer-inner-contrastive strategy is widely used for the evaluation process in subsequent works. Because the class activation sequence may only discover the most discriminative action parts, subsequent works aim at improving the quality of class activation sequences, via mining co-activity loss \cite{paul2018w}, learning multiple complementary class activation sequences CMCS \cite{liu2019completeness}, mining temporal structure \cite{yu2019temporal}. Recently, BaSNet \cite{lee2019background} achieves precise localization performance, i.e., mAP@$0.5$=$27.0$. Although above methods are able to improve the quality of class activation sequence to some extent, they essentially follow the pre-classification pipeline. Specifically, only local features of each point are used to learn the classifier, while lacking consideration to the global video features. As a result, ECM exceeds these methods with at least $2.1$ performance gains, under the guidance of the equivalence mechanism. Apart from above methods, we note that some recent methods \cite{gong2020learning, min2020adversarial, luo2020weakly, jain2020actionbytes} achieve high performance on THUMOS14 dataset. It should be noticed that these methods are somewhat complicated. For example, ActionBytes \cite{jain2020actionbytes} is designed to generate action proposals, which are then combined with W-TALC \cite{paul2018w} to localize action instances. EM-MIL \cite{luo2020weakly} requires UntrimmedNet \cite{wang2017untrimmednets} or W-TALC \cite{paul2018w} to classify action proposals. In contrast, ECM adopts a unified framework and a simple method, but achieves competitive performance. This demonstrates the efficiency of the equivalence classification mapping mechanism.

As for post-classification methods, STPN \cite{nguyen2018weakly} is an early and extensively used baseline with performance mAP@$0.5$=$16.9$. Because attention weights have essential influence on aggregated video-level features, learning accurate attention weights becomes a research focus for subsequent works \cite{yuan2019marginalized, nguyen2019weakly, zhai2020two, shi2020weakly}. Specifically, MAAN \cite{yuan2019marginalized} proposes marginalized average aggregation, and WSBM \cite{nguyen2019weakly} optimizes background attention weights. Recently, TSCN \cite{zhai2020two} and DGAM \cite{shi2020weakly} build similar high performance under the metric mAP@$0.5$, reaching $28.7$ and $28.8$, respectively.

Compared with existing pre-classification methods and post-classification methods, ECM delves into both local features at each point and global features for the complete video to learn the classification network. Under the metric mAP@$0.5$, ECM starts from simple pre-classification and post-classification baselines, and shows $2.1$ performance improvements over BaSNet \cite{lee2019background}, $0.3$ performance improvement over the strong competitor DGAM \cite{shi2020weakly}, without bells and whistles. This demonstrates the efficiency and superiority of the revealing equivalence mechanism.

Compared with fully supervised methods, on the one hand, ECM already exceeds some early methods, e.g., the one-stage method CDC \cite{shou2017cdc} and the two-stage method R-C3D \cite{xu2017r}. This demonstrates the efficiency of ECM for discovering action instances. On the other hand, the latest fully supervised methods achieve precise localization performance, e.g., PGCN \cite{zeng2019graph} achieves $49.1$ under metric mAP@$0.5$. This reminds us that the weakly supervised methods require further studies.

\begin{table}[htbp]
	\centering
	\caption{Ablation studies about network architecture and loss functions on THUMOS14 dataset, measured by mAP@$0.5$. "\CheckedBox" indicates not sharing classifiers between two streams.}
	\begin{tabular}{l|cccccc}
		\toprule
		Pre-Cls & \checkmark &       & \checkmark & \checkmark & \CheckedBox & \checkmark \\
		Post-Cls &       & \checkmark & \checkmark & \checkmark & \CheckedBox & \checkmark \\
		$\mathcal{L}_{a2c}$ &       &       &       & \checkmark & \checkmark & \checkmark \\
		$\mathcal{L}_{c2c}$ &       &       &       &       & \checkmark & \checkmark \\
		\midrule
		mAP   & 19.6  & 17.1  & 22.8  & 24.2  & 23.4  & 29.1 \\
		\bottomrule
	\end{tabular}%
	\label{tab:ablation-components}
\end{table}%

\subsection{Ablation studies}
\label{chapter-ablation-exp}

\begin{figure*}[htbp]
	\graphicspath{{figure/}}
	\centering
	\includegraphics[width=1\linewidth]{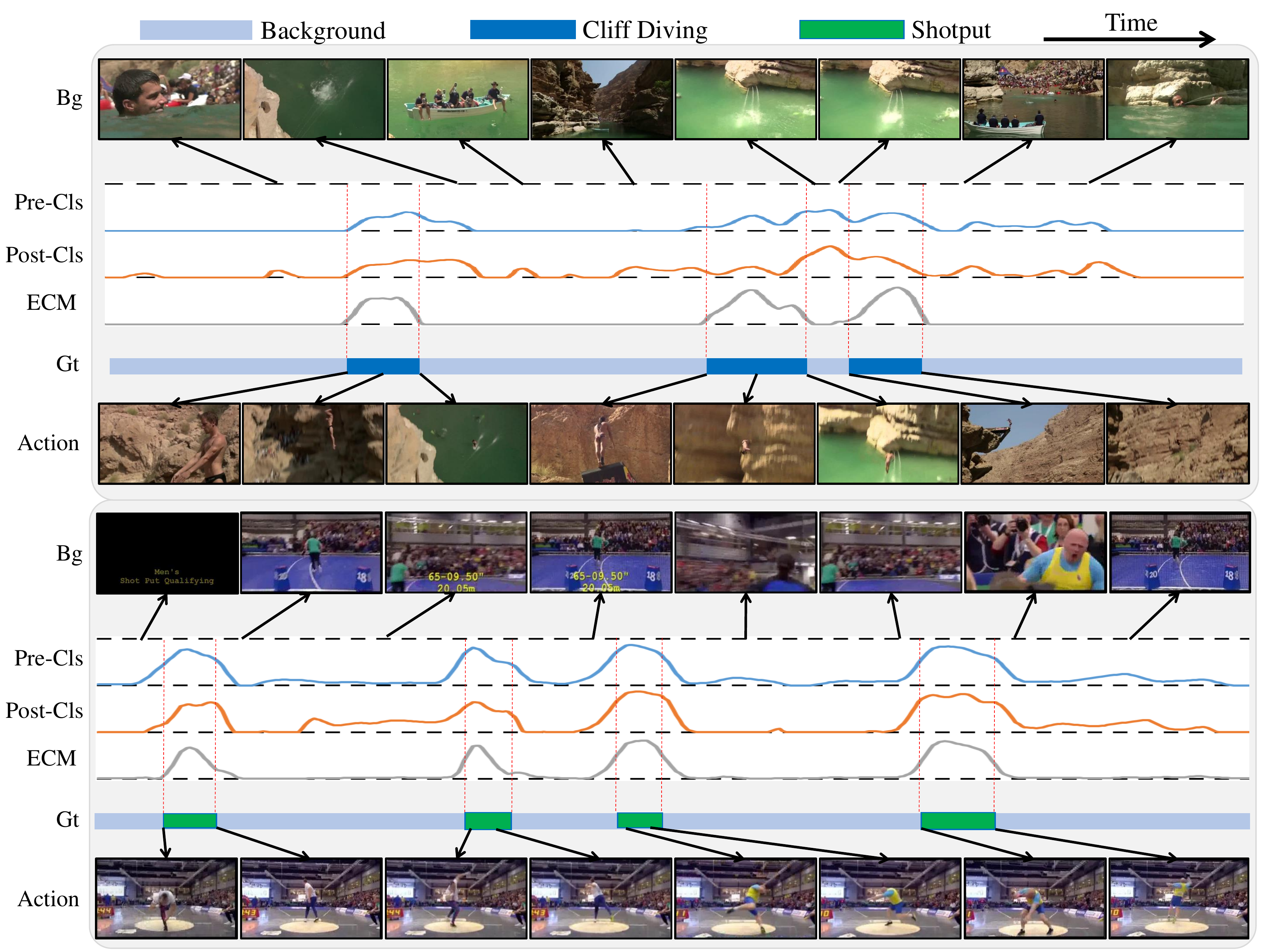}
	\caption{Qualitative results of the class activation sequences for pre-classification stream (Pre-Cls), post-classification stream (Post-Cls) and the proposed ECM.}
	\label{fig:visualization}
\end{figure*}

\textbf{Ablation studies about the equivalence mechanism.} We perform ablation studies on the THUMOS14 dataset and report the results in Table \ref{tab:ablation-components}. Under threshold $0.5$, the pre-classification stream and the post-classification stream can achieve $19.6$ and $17.1$, respectively. On this basis, we share the classifier between the pre-classification stream and the post-classification stream, transit the aggregation weights from the class activation sequence in the pre-classification stream. This direct and simple implementation of the equivalence mechanism reaches the performance of $22.8$. Based on this, the proposed aggregation-to-classification consistency training strategy can lift the performance to $24.2$. This supports our claim that constraining the consistency between aggregation weights and action presences. The aggregation-to-classification loss $\mathcal{L}_{a2c}$ guides the classifier to perceive both action presence and action absence, so as to precisely localize the starting time and ending time of action instances. Moreover, the complete ECM achieves $29.1$, under the cooperation of sharing the classifier, aggregation-to-classification loss $\mathcal{L}_{a2c}$ and classification-to-classification loss $\mathcal{L}_{c2c}$. Furthermore, to clearly verify the efficiency of the equivalence classification mapping mechanism, we remove classifier sharing from ECM and keep all other components unchanged. Under this setting, the variant model obtains $23.4$, which shows $5.7$ points inferior to the performance of ECM. This experiment demonstrates that classifier sharing (i.e., the equivalence mechanism) is the foundation of ECM, which effectively empowers the proposed equivalence-based components, i.e., the weight-transaction module, and the equivalent training strategy ($\mathcal{L}_{a2c}$ and $\mathcal{L}_{c2c}$). Beyond this, we verify the effectiveness of class-specific feature representation. We follow existing methods \cite{nguyen2018weakly, nguyen2019weakly, yuan2019marginalized} and learn class-agnostic aggregation weights for the post-classification stream, such a variant achieves the performance of $27.6$. In contrast, the $1.5$ performance improvements of ECM owes to rich feature representations from the class-specific attention weights.

\textbf{Differences with model ensemble.} In implementation, ECM maintains two streams with weight sharing, which practice the equivalence classification mapping mechanism. ECM is essentially different from model ensemble methods. Specifically, ensemble methods usually train multiple models in parallel, generate multiple heterogeneous results, and fuse results to get improvements. In contrast, ECM learns one classifier from a unified learning framework (with a  

\begin{table}[htbp]
	\centering
	\caption{Ablation studies about fusion localization results from the pre-classification stream and the post-classification stream.}
	\begin{tabular}{cccccccc}
		\toprule
		\multirow{2}[2]{*}{TH} & \multirow{2}[2]{*}{Pre-Cls} & \multirow{2}[2]{*}{Post-Cls} & \multicolumn{1}{c}{\multirow{2}[2]{*}{Merge}} & \multicolumn{4}{c}{Weighted sum $\lambda$} \\
		&       &       &       & 0.2   & 0.4   & 0.6   & 0.8 \\
		\midrule
		0.1   & 46.1  & 42.3  & 45.5  & 43.5  & 44.2  & 44.9  & 45.1  \\
		0.2   & 38.8  & 35.8  & 39.2  & 36.6  & 37.4  & 38.2  & 38.4  \\
		0.3   & 32.1  & 29.1  & 33.0  & 30.7  & 31.2  & 31.9  & 32.1  \\
		0.4   & 25.3  & 23.1  & 26.5  & 24.9  & 25.5  & 26.0  & 26.3  \\
		0.5   & 19.6  & 17.1  & \textbf{21.4} & 20.2  & 20.3  & 20.5  & 20.8  \\
		0.6   & 12.7  & 11.0  & 13.8  & 13.4  & 13.9  & 14.3  & 14.4  \\
		0.7   & 7.2   & 6.2   & 7.9   & 8.2   & 8.2   & 8.3   & 8.4  \\
		\bottomrule
	\end{tabular}%
	\label{tab:fuse-two-streams}%
\end{table}%

\begin{figure*}[htbp]
	\graphicspath{{figure/}}
	\centering
	\includegraphics[width=1\linewidth]{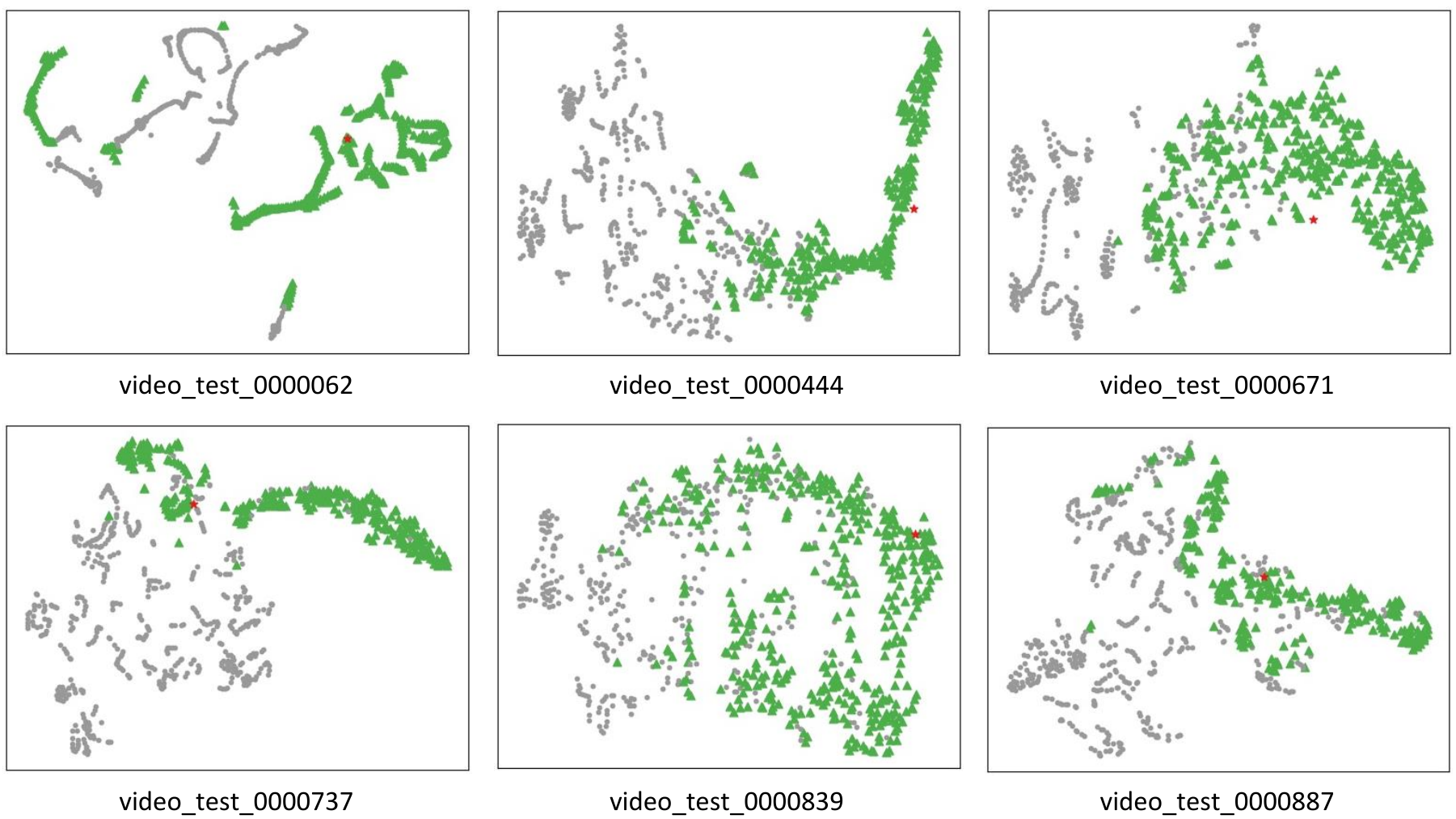}
	\caption{Mainfold distribution of point features and aggregated features. Background features, action features and the aggregated feature are shown in gray \textgray{$\bullet$}, green \textgreen{$\blacktriangle$} and red \textred{$\bigstar$}, respectively. Best viewed in zoom.}
	\label{fig:mainfold}
\end{figure*}

\noindent
two-stream architecture). In inference, ECM uses one model to predict one result rather than fusing results that are predicted by multiple models. Empirically, we perform ablation studies and report the results in Table \ref{tab:fuse-two-streams}. If we independently train the pre-classification network and the post-classification network, the performance is $19.6$ and $17.1$, respectively. There are two kinds of strategies to perform ensemble. The first one is action instances from two class activation sequences and merge the localization results. This strategy achieves $21.4$ under the metric mAP@$0.5$. The  

\begin{table}[htbp]
	\centering
	\caption{Ablation studies about frame accuracy on THUMOS14 dataset.}
	\begin{tabular}{cccc}
		\toprule
		& Pre-Classification & Post-Classification & ECM \\
		\midrule
		Frame Accuracy & 42.7\% & 39.0\% & 54.3\% \\
		\bottomrule
	\end{tabular}%
	\label{tab:frame-acc}%
\end{table}%

\noindent
second strategy is to fuse two class activation sequences and localize action instances from the fused sequence. Specifically, for class activation sequences from the pre-classification stream and the post-classification stream, we apply weight $\lambda$ and $1-\lambda$, respectively. We carry out experiments with different $\lambda$, i.e., $\lambda \in \{0.2$, $0.4$, $0.6$, $0.8\}$ and report the performance. The best performance is $20.8$ when $\lambda=0.8$. In general, the highest performance that model ensemble strategy can achieve is $21.4$. In Table \ref{tab:ablation-components}, a direct and simple implementation of the equivalence mechanism achieves $22.8$, and the proposed equivalence-based components (i.e., the weight-transaction module and equivalent training strategy) can further mine the equivalence classification mapping mechanism and improve the performance to $29.1$. The $8.7$ performance gap between model ensemble and ECM demonstrates the efficiency of the proposed equivalence mechanism, as well as the significant differences between ECM and model ensemble.

\textbf{Frame Accuracy.} In inference, the pre-classification method, the post-classification method and the proposed ECM method, all directly learn classifiers to predict the classification score for each temporal point. Following the \textit{frame accuracy} metric adopted by the action segment researches \cite{chang2019d3tw}, we measure the frame accuracy of the generated class activation sequences to validate the quality of the point-level classification. Similar to previous works \cite{chang2019d3tw}, we do not consider the background points in order to prevent the case that a method predicts all frames as background but still achieves high performance.

Given the obtained class activation sequences, we first predict the video-level classification scores and reject categories whose classification scores are lower than threshold $\tau=0.25$. Then, for the remaining responses, we perform max pooling and obtain the point-wise classification predictions. Finally, the frame-wise classification accuracy is reported in Table \ref{tab:frame-acc}. It can be found that the pre-classification method performs somewhat better than the post-classification method, with a margin of $3.7\%$. The proposed ECM brings obvious improvements, with a margin of $11.6\%$. The results demonstrate that the proposed equivalent classification mapping mechanism can learn a high-quality classifier and can accurately discover action snippets in the untrimmed videos.

\subsection{Qualitative Results}

\textbf{Visualization of action localizations.} We qualitatively visualize the class activation sequences in Figure \ref{fig:visualization}. When only using the pre-classification stream or the post-classification stream, the generated class activation sequences cannot distinguish two adjacent action instances (see the first case), or shows high responses for a part of background points (see the second case). This may lead to confusing localization results. In contrast, the proposed ECM method precisely shows high responses for action instances and confidently suppresses the response of backgrounds, leading to accurate localization results.

\textbf{Visualization of feature distribution.} At the beginning of Section \ref{chaper-method-network}, it is proven that the aggregated video-level features lie in the same feature space with point-level features. In Figure \ref{fig:mainfold}, we use t-SNE \cite{maaten2008visualizing} to visualize the feature distribution for point features and aggregated features. It is clear that action features and background features lie in different mainfolds. Besides, the aggregated video-level features lie in the same mainfold with the action features. This visualization supports our claim that the aggregated video-level features possess similar property with action features. This builds a solid foundation for the equivalence classification mapping mechanism.

\section{Conclusion}
In this paper, we propose the equivalent classification mapping mechanism to weakly supervised temporal action localization task. Specifically, ECM starts from both the pre-classification stream and the post-classification stream to simultaneously learn one classification mapping function. Assisted with the weight-transition module and equivalent training strategy, ECM achieves accurate action localization performance on three benchmarks. Considering ECM is simple to implement and achieves good performance without bells and whistles, it can serve as a solid baseline for subsequent researches. Furthermore, it is a promising direction that applying the inspiration of ECM to similar research areas, e.g., weakly supervised object localization \cite{zhou2016learning}, segmentation \cite{zhang2018spftn}, detection \cite{zhang2018learning, zhang2019leveraging, zhang2020discriminant} .

%\bibliographystyle{IEEEtran}
%\bibliography{egbib}
% Generated by IEEEtran.bst, version: 1.13 (2008/09/30)

% Can use something like this to put references on a page
% by themselves when using endfloat and the captionsoff option.
\ifCLASSOPTIONcaptionsoff
  \newpage
\fi

%\begin{thebibliography}{1}
%
%\bibitem{IEEEhowto:kopka}
%H.~Kopka and P.~W. Daly, \emph{A Guide to \LaTeX}, 3rd~ed.\hskip 1em plus
%  0.5em minus 0.4em\relax Harlow, England: Addison-Wesley, 1999.
%
%\end{thebibliography}

\begin{IEEEbiography}[{\includegraphics[width=1in,height=1.25in,clip,keepaspectratio]{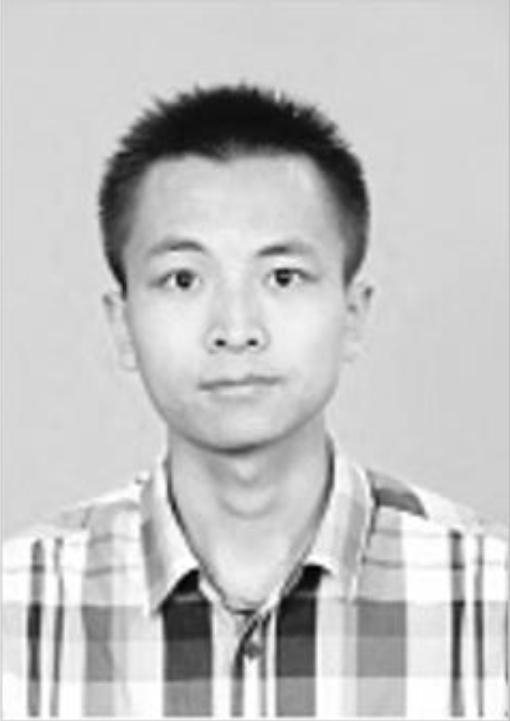}}]{Tao Zhao} received his M.S. degree from Northwestern Polytechnical University, Xi'an, China, in 2018. He is currently a Ph.D. candidate in the School of Automation at Northwestern Polytechnical University. His research interests include video temporal action localization and weakly supervised learning.
\end{IEEEbiography}

\begin{IEEEbiography}[{\includegraphics[width=1in,height=1.25in,clip,keepaspectratio]{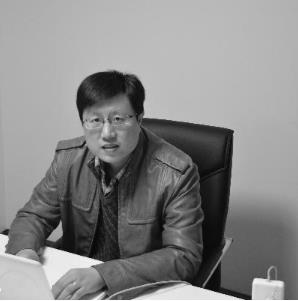}}]{Junwei Han}
	is currently a Professor in the School of Automation, Northwestern Polytechnical University. His research interests include computer vision, pattern recognition, remote sensing image analysis, and brain imaging analysis. He has published more than 70 papers in top journals such as IEEE TPAMI, TNNLS, IJCV, and more than 30 papers in top conferences such as CVPR, ICCV, MICCAI, and IJCAI. He is an Associate Editor for several journals such as IEEE TNNLS and IEEE TMM.
\end{IEEEbiography}

\begin{IEEEbiography}[{\includegraphics[width=1in,height=1.25in,clip,keepaspectratio]{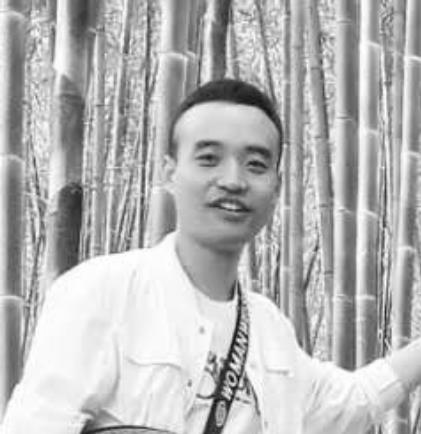}}]{Le Yang} received his B.E. degree from Northwestern Polytechnical University, Xi'an, China, in 2016. He is currently a Ph.D. candidate in the School of Automation at Northwestern Polytechnical University. His research interests include video action localization and video object segmentation.
\end{IEEEbiography}

\begin{IEEEbiography}[{\includegraphics[width=1in,height=1.25in,clip,keepaspectratio]{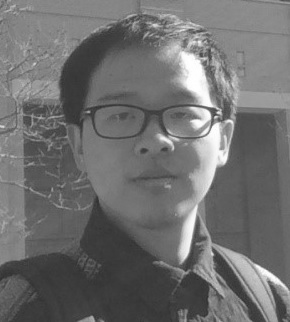}}]{Dingwen Zhang} received his Ph.D. degree from the Northwestern Polytechnical University, Xi'an, China, in 2018. He is currently an associate professor in the School of Machine-Electronical Engineering, Xidian University. From 2015 to 2017, he was a visiting scholar at the Robotic Institute, Carnegie Mellon University. His research interests include computer vision and multimedia processing, especially on saliency detection, video object segmentation, and weakly supervised learning.
\end{IEEEbiography}

% that's all folks
\end{document}